\documentclass[11pt,a4paper]{article}
\usepackage{naaclhlt2018}
\usepackage{times}
\usepackage{url}
\usepackage{latexsym}
\usepackage{amsmath}
\usepackage{multirow}
\usepackage{rotating}
\usepackage{amssymb}
\usepackage{graphicx}
\usepackage[small]{caption}
\usepackage{xspace}
\usepackage[utf8]{inputenc}
\usepackage{microtype}
\usepackage{algorithm}
\usepackage{booktabs}
\usepackage[noend]{algpseudocode}
\usepackage{titlesec}

\usepackage{inconsolata}

\aclfinalcopy

\newcommand{\marmot}{\sw{MarMoT}\xspace}
\newcommand{\cistern}{{\small \url{http://cistern.cis.lmu.de/lemming}\xspace}}

\hypersetup{pdfauthor={Thomas Müller, Ryan Cotterell, Alexander Fraser, Hinrich Schütze}, pdftitle={Joint Lemmatization and Morphological Tagging with Lemming}, pdfkeywords={morphology, NLP, morphological analysis, lemma}}

\newcommand{\FirstPage}{2268}
\newcommand{\LastPage}{2274}

\newcommand{\citeinfo}[2]{
  \AddToShipoutPicture*{%
    \setlength{\unitlength}{1mm}
    \put(105,17){\makebox(0,0){\footnotesize {\em Proceedings of the 2015 Conference on Empirical Methods in Natural Language Processing}, pages #1--#2}}
     \put(105,13){\makebox(0,0){\footnotesize Lisbon, Portugal, 17-21 September 2015. \textcopyright 2015 Association for Computational Linguistics}}
  }
}

\citeinfo{\FirstPage}{\LastPage}

\def\sigsym{\mbox{$+$}}
\def\kigsym{\mbox{$\times$}}
\def\tigsym{\rlap{\sigsym}\kigsym}

\def\sig{\mbox{$^{+}$}}
\def\kig{\mbox{$^{\times}$}}
\def\tig{\mbox{\rlap{\sig}{\kig}}}
\def\nig{\mbox{\phantom{\sig}}}

\def\figref#1{Figure~\ref{fig:#1}}
\def\figlabel#1{\label{fig:#1}\label{p:#1}}

\def\tabref#1{Table~\ref{tab:#1}}
\def\tablabel#1{\label{tab:#1}}

\def\secref#1{Section~\ref{sec:#1}}
\def\seclabel#1{\label{sec:#1}\label{p:#1}}
\def\eqref#1{Eq.~\ref{eqn:#1}}

\newcommand{\all}{\multicolumn{1}{l}{\scriptsize all}}
\newcommand{\unk}{\multicolumn{1}{l}{\scriptsize unk}}

\newcommand{\bftab}{\fontseries{b}\selectfont}
\newcommand{\tb}[1]{{\bftab #1}}
\newcommand{\ul}[1]{\underline{#1}}
\newcommand{\noul}[1]{#1}
\newcommand{\score}[1]{{\tiny #1}}

\newcommand{\cdotline}[1]{}%

\newcommand{\tree}{\operatorname{tree}}

\renewcommand{\vec}{\boldsymbol}   %

\newcommand{\NA}{\multicolumn{1}{c}{NA}}

\newcommand{\vf}{{\vec{f}}}
\newcommand{\vg}{{\vec{g}}}

\newcommand{\vtheta}{{\vec{\theta}}}
\newcommand{\vlambda}{{\vec{\lambda}}}

\newcommand{\software}[1]{\textsc{#1}}
\newcommand{\sw}[1]{\software{#1}}
\newcommand{\word}[1]{{\em #1}}
\newcommand{\gloss}[1]{``#1''}

\usepackage[disable]{todonotes}

\newcommand{\Alex}[1]{\todo[inline,color=purple!40,author=Alex]{#1}}

\title{Joint Lemmatization and Morphological Tagging with \software{Lemming}}

\author{Thomas M{\"u}ller$^{1}$\quad Ryan Cotterell$^{1,2}$ \quad   Alexander Fraser$^{1}$  \quad  Hinrich Sch{\"u}tze$^{1}$ \\
  Center for Information and Language Processing$^1$, University of Munich, Germany  \\
Department of Computer Science$^2$, Johns Hopkins University, USA \\
\texttt{\href{mailto:muellets@cis.lmu.de}{muellets@cis.lmu.de}}\quad \texttt{\href{mailto:ryan.cotterell@jhu.edu}{ryan.cotterell@jhu.edu}}}

\date{}

\long\def\eat#1{\ignorespaces}
\newcommand{\enotesoff}{\long\gdef\enote##1##2{}}

\enotesoff

\newcommand{\camera}[1]{#1}

\begin{document}
\maketitle

\begin{abstract}
We present \software{Lemming}, a modular log-linear model that
jointly models lemmatization and tagging and supports the integration
of arbitrary global features. It is trainable on corpora annotated
with gold standard tags and lemmata and does not rely on morphological
dictionaries or analyzers. \software{Lemming} sets the new
state of the art in token-based statistical lemmatization on six
languages; e.g., for Czech lemmatization, we reduce the error by 60\%,
from 4.05 to 1.58.  We also give empirical evidence that jointly
modeling morphological tags and lemmata is mutually beneficial.
\end{abstract}

\section{Introduction}
Lemmatization is important for many NLP tasks, including
parsing \cite{bjorkelund2010,seddah2010} and machine translation
\cite{fraser2012}. Lemmata are required whenever we want to
map words to lexical resources and establish the relation
between inflected forms,
particularly critical 
for
morphologically rich languages to address the sparsity of
unlemmatized forms. 
This strongly motivates work 
on 
language-independent token-based lemmatization, but
until now there 
has been
little work \cite{chrupala2008learning}.

Many regular transformations can
be described by simple replacement rules,
but lemmatization of unknown words 
requires more than this.
For instance
the Spanish
paradigms for verbs ending in \word{ir}
and \word{er} share the same 3rd person plural ending \word{en}; this makes it hard to decide
which paradigm a form belongs to.\footnote{Compare \word{admiten} \gloss{they admit} $\rightarrow$
\word{admitir} \gloss{to admit}, but \word{deben} \gloss{they must} $\rightarrow$
\word{deber} \gloss{to must}.}
Solving these kinds of problems requires
global features on the lemma.
Global features of this kind were not
supported by
previous work
\cite{dreyer2008latent,chrupala2006simple,toutanova2009global,cotterell2014stochastic}.

There is a strong mutual dependency between (i) 
lemmatization of a form in context and (ii)
disambiguating its part-of-speech (POS) and morphological attributes.
Attributes often disambiguate the lemma of a form, which explains
why many NLP systems \cite{manning2014,padro2012} apply a pipeline approach of
tagging followed by lemmatization.
Conversely, knowing the lemma of a form is often  beneficial
for tagging,
for instance
in the presence of syncretism;
e.g., 
\Alex{Few German nouns mark gender even in singular, so this is a bit confusing. I tried to fix it by referring to noun phrases and the gender of the head noun. But the point actually being made here is that we can infer the gender from noun phrase instances where it is clear, and using the lemma allows us to link these with the plural and other unclear noun phrases.}
since German plural noun phrases do not mark gender,
it is important to know the lemma (singular form) 
to correctly tag gender on the noun.

We make the following contributions.
(i)
We present the first joint  log-linear model of morphological
analysis and lemmatization that operates
at the {\em token} level and is also able to lemmatize
unknown forms; and release it as open-source (\cistern).
It is trainable
on corpora annotated with gold standard tags and lemmata.
Unlike other work 
(e.g., 
\citet{smith2005}) 
it
does not rely on morphological dictionaries or analyzers.
(ii)
We describe a log-linear model for lemmatization that can easily be
incorporated into other models and supports
arbitrary global features on the lemma.
(iii)
We set the new state of the art in token-based statistical
lemmatization on six languages (English, German, Czech,
Hungarian, Latin and Spanish).
(iv)
We experimentally show
that jointly modeling morphological
tags and lemmata is mutually beneficial
\Alex{It seems wrong here that there is no mention of improved ``tag'' results?}
and yields significant improvements in joint (tag+lemma) accuracy
for four out of six languages;
e.g., Czech lemma errors are reduced by $>$37\% and tag+lemma errors by $>$6\%.\looseness=-1

\section{Log-Linear Lemmatization}\seclabel{lemmatization}

\newcite{chrupala2006simple}
formalizes
lemmatization as a classification task through the
deterministic pre-extraction of edit operations transforming
forms into lemmata.
Our lemmatization
model is in this vein,
but allows
the addition of external lexical
information, e.g., whether the candidate lemma is in a dictionary.
Formally, lemmatization is a string-to-string
transduction task.
Given an alphabet $\Sigma$,
it maps an inflected form $w \in \Sigma^*$ to its lemma $l \in
\Sigma^*$ given its morphological attributes $m$. We model this process by a
log-linear model:
\begin{equation}p(l \mid w, m) \propto h_w(l) \cdot \exp \left(\vf(l,w, m)^{\top} \vtheta
\right),
\end{equation}
where $\vf$ 
represents
hand-crafted feature 
functions,
$\vtheta$ is a
weight vector, and
$h_w:\Sigma^* \rightarrow \{0,1\}$
determines the support of the distribution,
i.e., the set of candidates with non-zero
probability.
\begin{figure}
   \centering
    \includegraphics[width=0.40\textwidth]{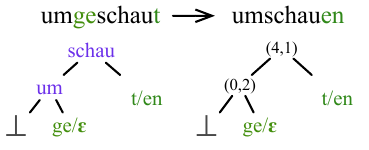}
\caption{Edit tree for the inflected form \word{umgeschaut} \gloss{looked around}
and its lemma \word{umschauen} \gloss{to look around}.
The right tree is the actual edit tree we use in our model,
the left tree visualizes what each node corresponds to.
The root
node
stores the length of the prefix \word{umge} (4) and the
suffix \word{t} (1).}
\figlabel{edit-tree}
\vspace{-0.25cm}
\end{figure}
\subsection{Candidate selection}
A proper choice of the support function $h(\cdot)$ is crucial
to the success of the model -- too permissive a function
and the computational cost will build up, too restrictive
and the correct lemma may receive no probability mass.
Following \newcite{chrupala2008towards}, we
define $h(\cdot)$ through a deterministic pre-extraction of
\emph{edit trees}.
To extract an edit tree $e$ for a pair form-lemma $\langle w,l \rangle$,
we first find the longest common substring (LCS) \cite{gusfield1997}
between them and then recursively model the prefix and suffix
pairs of the LCS.
When
no LCS can be found the 
string pair is represented as
a substitution operation transforming the first 
string to the second.
The 
resulting edit
tree does not encode the LCSs but only the length of their prefixes
and suffixes and the substitution nodes
(cf.\ \figref{edit-tree}); e.g.,
the same tree  transforms
\word{worked}
into
\word{work} and \word{touched} into \word{touch}.\looseness=-1

As a preprocessing step, we extract all edit trees that can be used for more
than one pair $\langle w,l \rangle$.
To generate the candidates of a
word-form,
we
apply all edit trees and also add all lemmata this form was seen with
in the training set
(note that only a small subset of
the
edit trees is
applicable for any given form because most require
incompatible substitution operations).

\subsection{Features}
Our
novel
formalization lets us  combine
a wide variety of
features that have been used in
different
previous
models.
All features are extracted given
a form-lemma pair $\langle w,l \rangle$ created with an
edit tree $e$.

We use the following three \emph{edit tree features} of \newcite{chrupala2008towards}.
(i) The edit tree $e$.
(ii) The pair $\langle e,w \rangle$.
This feature is crucial for the model to memorize irregular forms,
e.g., the lemma of \word{was} is \word{be}.
(iii) For each form affix (of maximum length 10):
its conjunction with $e$.
These features are
useful
in learning orthographic and phonological regularities, e.g.,
the lemma of \word{signalling} is \word{signal}, not
\word{signall}.\looseness=-1

\begin{figure}
   \centering
    \includegraphics[width=0.45\textwidth]{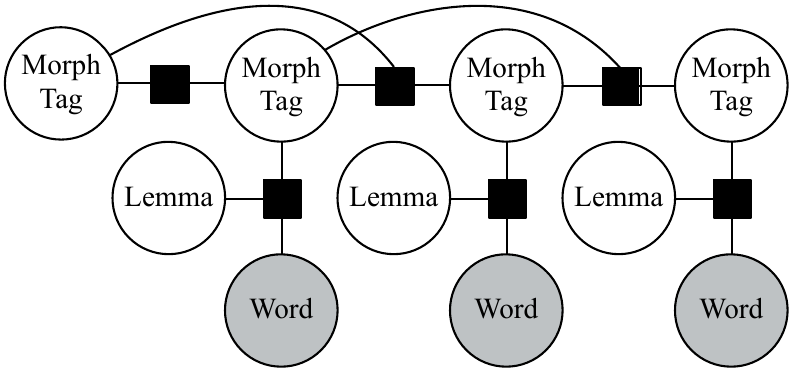}
 \caption{Our  model
is a 2nd-order linear-chain CRF augmented
to predict lemmata.
We heavily prune
our model and can easily exploit higher-order ($>$2) tag dependencies.}
\vspace{-0.25cm}
\label{fig:factor-graph}
\end{figure}

We define the following \emph{alignment features.}
Similar to \newcite{toutanova2009global} (TC),
we define an alignment between $w$ and $l$.  Our alignments can be
read from an edit tree by aligning the characters in LCS nodes
character by character and characters in substitution nodes
block-wise. Thus the alignment of \word{umgeschaut} - \word{umschauen}
is: u-u, m-m, ge-$\epsilon$, s-s, c-c, h-h, a-a, u-u, t-en.  Each
alignment pair constitutes a feature in our model.  These features
allow the model to learn that the substitution \word{t}/\word{en} is likely in
German. We also concatenate each alignment pair with its form and lemma
character context (of up to length 6) to learn, e.g.,  that
\word{ge} is often deleted after  \word{um}.

We define two simple
\emph{lemma features}.
(i) 
We use the lemma itself as a feature, allowing
us to learn
which
lemmata are common in the language.
(ii) Prefixes and suffixes of the lemma (of maximum length 10).
This feature allows us to learn that the typical endings
of Spanish verbs are \word{ir}, \word{er}, \word{ar}.

We also use two \emph{dictionary features} (on lemmata):
Whether $l$ occurs $> 5$ times in Wikipedia
and whether it occurs in the dictionary \software{Aspell}.\footnote{\url{ftp://ftp.gnu.org/gnu/aspell/dict}}
\camera{We use a similar feature for different
capitalization variants of the lemma
(lowercase, first letter uppercase, all uppercase, mixed). 
This differentiation is important for German, where nouns are
capitalized and \word{en} is both a noun plural marker and a
frequent verb ending. Ignoring
capitalization would thus lead to confusion.}

\emph{POS \& morphological attributes.}
For each feature 
listed previously,
we create a conjunction with the POS
and each morphological attribute.\footnote{Example: for the Spanish noun \word{medidas} \gloss{measures}
with attributes \textsc{Noun}, \textsc{Common}, \textsc{Plural} and \textsc{Feminine},
we conjoin each feature above with \textsc{Noun}, \textsc{Noun+Common}, \textsc{Noun+Plural} and \textsc{Noun+Feminine}.}

\section{Joint Tagging and Lemmatization}\seclabel{joint}

We model the sequence of morphological tags using \marmot \cite{mueller2013}, a pruned higher-order CRF.
This model avoids the exponential runtime of higher-order models
by employing a pruning strategy.
Its feature set consists of standard tagging features: the current word, its affixes and shape (capitalization, digits, hyphens) and the
immediate lexical context.
We combine lemmatization and  higher-order CRF components
in a tree-structured CRF. Given a
sequence of forms $\vec{w}$ with lemmata $\vec{l}$ and morphological+POS
tags $\vec{m}$, we define a globally normalized model:
\begin{align}
  p(\vec{l},\vec{m} \mid &\vec{w}) \propto   \prod_{i=1}^{|\vec{w}|} h_{w_i}(l_i)  \exp\Big( \vf(l_i,w_i,m_i)^{\top}\vtheta \nonumber \\ 
 & +  \vg(m_i,m_{i-1},m_{i-2}, \vec{w}, i)^{\top}\vlambda\Big),
\end{align}
where $\vf$ and $\vg$ are the features associated with
lemma and tag cliques respectively and $\vtheta$ and $\vlambda$ are weight vectors.
The graphical model is shown
in Figure \ref{fig:factor-graph}.
We perform inference with belief propagation \cite{pearl1988}
and estimate the
parameters with SGD \cite{tsuruoka2009stochastic}.
We greatly improved
the results of the joint model by initializing it
with the parameters of
a pretrained tagging model.\looseness=-1

\begin{table*} 
\renewcommand\thetable{2}
\centering
\small
\setlength{\tabcolsep}{2pt}
\begin{tabular}{r@{\hskip 8pt}l@{\hskip 8pt}ll@{\hskip 4pt} rr rr rr rr rr rrl} \toprule
\multicolumn{4}{c}{} & \multicolumn{2}{c}{cs} & \multicolumn{2}{c}{de} & \multicolumn{2}{c}{en} & \multicolumn{2}{c}{es} & \multicolumn{2}{c}{hu} & \multicolumn{2}{c}{la} \\
&&&                      & \all & \unk & \all & \unk & \all & \unk & \all & \unk & \all & \unk & \all & \unk \\
\hline
\score{1}& \multicolumn{2}{c}{\scriptsize \marmot}& \score{tag} & \noul{89.75}\nig & \noul{76.83}\nig & \noul{82.81}\nig & \noul{61.60}\nig & \noul{\tb{96.45}}\nig & \noul{\tb{90.68}}\nig & \noul{97.05}\nig & \noul{90.07}\nig & \noul{93.64}\nig & \noul{84.65}\nig & \noul{82.37}\nig & \noul{53.73}\nig \\\hline
\score{2}&\multirow{2}{*}{\begin{sideways} \scriptsize JCK\end{sideways}}&& \score{lemma} & \noul{95.95}\nig & \noul{81.28}\nig & \noul{96.63}\nig & 85.84\nig & \noul{99.08}\nig & \noul{94.28}\nig & 97.69\nig & 87.19\nig & \noul{96.69}\nig & \noul{88.66}\nig & \noul{90.79}\nig & \noul{58.23}\nig \\\cdotline{5-16}
\score{3}&& & \score{\score{tag}+\score{lemma}} & \noul{87.85}\nig & \noul{67.00}\nig & \noul{81.60}\nig & \noul{55.97}\nig & \noul{96.17}\nig & \noul{87.32}\nig & \noul{95.44}\nig & 80.62\nig & \noul{92.15}\nig & \noul{78.89}\nig & \noul{79.51}\nig & \noul{39.07}\nig \\
\midrule
\score{4}&
\multirow{4}{*}{\begin{sideways} \scriptsize \sw{Lemming-P}\end{sideways}}& \multirow{2}{*}{\begin{sideways}{\scriptsize +dict}\end{sideways}} & \score{lemma} & 97.46\nig & 89.14\nig & 97.70\nig & 91.27\nig & \tb{99.21}\nig & \tb{95.59}\nig & 98.48\nig & 92.98\nig & 97.53\nig & 92.10\nig & 93.07\nig & 69.83\nig \\\cdotline{5-16}
\score{5}&& & \score{\score{tag}+\score{lemma}} & 88.86\nig & 72.51\nig & 82.27\nig & 59.42\nig & \tb{96.27}\nig & \tb{88.49}\nig & 96.12\nig & 85.80\nig & 92.59\nig & 80.77\nig & 80.49\nig & 44.26\nig \\\cline{3-16}
\score{6}& &\multirow{2}{*}{\begin{sideways}{\scriptsize +mrph}\end{sideways}} & \score{lemma} & 97.29\nig & 88.98\nig & 97.51\nig & 90.85\nig &\NA & \NA & 98.68\nig & 94.32\nig & 97.53\nig & 92.15\nig & 92.54\nig & 67.81\nig \\\cdotline{5-16}
\score{7}&& & \score{\score{tag}+\score{lemma}} & 89.23\nig & 74.24\nig & 82.49\nig & 60.42\nig &\NA &\NA & 96.35\nig & 87.25\nig & 93.11\nig & 82.56\nig & 80.67\nig & 45.21\nig \\\hline
\score{8}&\multirow{6}{*}{\begin{sideways} \scriptsize \sw{Lemming-J}\end{sideways}}&\multirow{3}{*}{\begin{sideways}{\scriptsize +dict}\end{sideways}} & \score{tag} & \tb{90.34}\sig & 78.47\nig & \tb{83.10}\sig & 62.36\nig & 96.32\nig & 89.70\nig & 97.11\nig & 90.13\nig & 93.64\nig & 84.78\nig & 82.89\nig & 54.69\nig \\\cdotline{5-16}
\score{9}&&& \score{lemma} & 98.27\nig & 92.67\nig & \tb{98.10}\sig & 92.79\nig & \tb{99.21}\nig & 95.23\nig & 98.67\nig & 94.07\nig & 98.02\nig & 94.15\nig & \tb{95.58}\sig & \tb{81.74}\sig \\\cdotline{5-16}
\score{10}&& & \score{\score{tag}+\score{lemma}} & 89.69\nig & 75.44\nig & 82.64\nig & 60.49\nig & 96.17\nig & 87.87\nig & 96.23\nig & 86.19\nig & 92.84\nig & 81.89\nig & 81.92\nig & 49.97\nig \\\cline{4-16}
\score{11}&& \multirow{3}{*}{\begin{sideways}{\scriptsize +mrph}\end{sideways}} & \score{tag}\nig & 90.20\nig & \tb{79.72}\tig & \tb{83.10}\sig & \tb{63.10}\tig & \NA & \NA & \tb{97.16}\nig & \tb{90.66}\nig & \tb{93.67}\nig & \tb{85.12}\nig & \tb{83.49}\tig & \tb{58.76}\tig \\\cdotline{5-16}
\score{12}&& & \score{lemma} & \tb{98.42}\tig & \tb{93.46}\tig & \tb{98.10}\sig & \tb{93.02}\sig & \NA & \NA & \tb{98.78}\tig & \tb{94.86}\tig & \tb{98.08}\sig & \tb{94.26}\sig & 95.36\nig & 80.94\nig \\\cdotline{5-16}
\score{13}&&& \score{\score{tag}+\score{lemma}} & \tb{89.90}\tig & \tb{78.34}\tig & \tb{82.84}\tig & \tb{62.10}\tig & \NA & \NA & \tb{96.41}\kig & \tb{87.47}\kig & \tb{93.40}\tig & \tb{84.15}\tig & \tb{82.57}\sig & \tb{54.63}\sig \\ \bottomrule
\end{tabular}
\caption{\emph{Test} results for 
\sw{lemming-j}, the joint
  model, and pipelines (lines 2--7) of \marmot\ and (i) JCK
  and (ii)
  \sw{lemming-p}. In each cell, overall token
  accuracy is left (all), accuracy on unknown forms is right (unk).
Standalone \marmot\  tagging accuracy (line 1) is not repeated for
pipelines (lines 2--7). The best numbers are bold.
\sw{lemming-j} models significantly better than \sw{lemming-p} (\sigsym), or \sw{lemming} models not using morphology (\emph{+dict}) (\kigsym) or both (\tigsym) are marked.
More baseline numbers in the appendix (Table A2).}
\tablabel{joint}
\end{table*}

\enote{hs}{i just realized that i don't understand
  ``POS-only''. is this supposed to mean ``not using
  morphological attributes?''}

\section{Related Work}\label{sec:related-work}

\Alex{The Morfette discussion is hard to follow. I think we should explicitly say there are two versions, one which uses edit operations and one which uses edit trees?}
\Alex{Include a version number for the Morfette we are using (presumably there will be new versions in the future...)}
\Alex{This discusson doesn't seem to clearly say that Morfette is a pipeline rather than a joint model, isn't that a key point?}
In functionality, our system resembles \software{Morfette}
\cite{chrupala2008learning}, which generates lemma candidates
by extracting edit operation sequences between lemmata and 
surface
forms \cite{chrupala2006simple},
and
then trains two maximum entropy
Markov models \cite{ratnaparkhi1996maximum} for morphological tagging
and lemmatization, which are queried using a beam search decoder. 

In our experiments we use the latest version\footnote{\url{https://github.com/gchrupala/morfette/commit/ca886556916b6cc1e808db4d32daf720664d17d6}} 
of \software{Morfette}.
This version is based on structured
perceptron learning \cite{collins02discriminativetraining} 
and edit trees \cite{chrupala2008towards}.
Models similar to \software{Morfette} include 
those of
\newcite{bjorkelund2010} and \newcite{gesmundo2012}
\Alex{IMO this is a bit weird. Why do we care about generation here, we haven't discussed it previously? Also, the wording is poor.}
and have also been used for generation \cite{duvsek2013}.
\Alex{It is isn't clear what is similar and what is different WRT Wicentowski}
\newcite{wicentowski02modelingand} similarly
treats
lemmatization as classification over a deterministically chosen
candidate set, but uses distributional information extracted from
large corpora as a key source of information.

\newcite{toutanova2009global}'s
joint morphological analyzer predicts the set of possible
lemmata and \emph{coarse-grained POS} for a word \emph{type}.
This is different from our problem of
lemmatization and \emph{fine-grained} morphological tagging of
\emph{tokens in context}. Despite
the superficial similarity of the two problems, \emph{direct
comparison is not possible}. 
TC's model
is best thought of as inducing a tagging dictionary
for OOV types, mapping them to a set of tag and lemma pairs,
whereas \software{Lemming}
is a token-level, context-based morphological tagger.

We do, however, use TC's model of lemmatization,
a string-to-string transduction model based on
\newcite{jiampojamarn08} (JCK),
as a
stand-alone baseline.
Our tagging-in-context model is faced with higher
complexity of learning and inference since it addresses a
more difficult task; thus, while we
could in principle use JCK as a replacement for our candidate selection,
the edit tree approach -- which has
 high coverage  at a low average number
of lemma candidates (cf. \secref{experiments}) -- allows us to
train and apply \software{Lemming} efficiently.

\newcite{smith2005} proposed a log-linear model for the context-based
disambiguation of a morphological dictionary. This
has the effect of joint tagging, morphological segmentation
and lemmatization, 
but, critically, is limited to the entries in the morphological dictionary (without
which the approach cannot be used), causing problems of recall.
In contrast, \software{Lemming} can analyze
any word, including OOVs, and only requires 
the same training corpus as a generic tagger
(containing tags and lemmata),
a
resource that is 
available
for many languages.\looseness=-1

\section{Experiments}
\seclabel{experiments}

\paragraph{Datasets.}
We present experiments on the joint task
of
lemmatization and tagging in
six diverse languages: English, German, Czech,
Hungarian, Latin and Spanish.
We use the same data sets as in \newcite{mueller2015}, but do not use the out-of-domain test sets.
The English data is from the Penn Treebank \cite{marcus1993building}, Latin from PROIEL \cite{haug2008creating}, German and Hungarian from SPMRL 2013
\cite{seddah2013overview}, and Spanish and Czech  from CoNLL 2009  \cite{hajivc2009conll}.
For German, Hungarian, Spanish and Czech we use the splits from the
shared tasks; for English the split from SANCL \cite{petrov2012}; and
for Latin a 8/1/1 split into \emph{train/dev/test}.
\camera{For all
languages we limit our training data to the first 100,000 tokens.}
Dataset statistics can be found in Table A4 of the appendix.
\camera{The lemma of Spanish  \word{se} is set to be consistent.}

\paragraph{Baselines.}
\camera{We compare our model to three
baselines.}
(i) \software{Morfette}
(see \secref{related-work}).
(ii)
\software{simple},
a  system that for each form-POS pair, returns the most frequent
lemma in the training data or the form if the pair is unknown. (iii)
JCK, our reimplementation of \newcite{jiampojamarn08}.
Recall that JCK is TC's lemmatization model and that
the full TC model is a type-based model that cannot be applied
to our task. 
As JCK struggles to memorize irregulars, we only use it for unknown
form-POS pairs and use \software{simple}
otherwise. %
For aligning the training data we use the
edit-tree-based alignment described in the feature section.  We only
use output alphabet symbols that are used for 
$\geq 5$ 
form-lemma
pairs and also add a special output symbol that indicates that the
aligned input should simply be copied.
We train the model using a structured averaged perceptron and
stop after 10 training iterations.
In preliminary experiments we
found type-based training to outperform token-based training.
This is understandable as we only apply our model to unseen form-POS pairs.
The feature set is an exact reimplementation of \cite{jiampojamarn08},
it consists of input-output pairs and their character context in a
window of 6.

\paragraph{Results.}
Our candidate selection strategy
results in an average number of lemma candidates between $7$ (Hungarian) and
$91$ (Czech) and a coverage of the correct lemma on \emph{dev} of $>$99.4 (except 98.4 for Latin).\camera{\footnote{Note that our definition of
  lemmatization accuracy and unknown forms ignores capitalization.}}
We first compare the baselines to \software{Lemming-P}, a
pipeline based on \secref{lemmatization}, that  lemmatizes a word
given a predicted tag and is trained using L-BFGS \cite{liu1989limited}.
We use the implementation of \software{Mallet} \cite{mccallum2002}.
For these experiments we train all models on gold
attributes and test on
attributes
predicted by \software{Morfette}.
\software{Morfette}'s lemmatizer can only be used with its own tags. We thus use \software{Morfette} tags to have a uniform setup, which isolates the effects of the different taggers. Numbers for \marmot tags are in the appendix (Table A1).
For the initial experiments, we only use POS and ignore
additional morphological attributes.
We use
different feature sets to illustrate the utility of our templates.

The first model uses the \emph{edit tree features} (edittree).
\tabref{features} shows that this version of \software{Lemming}
outperforms the baselines on half of the languages.\footnote{Unknown word accuracies in the appendix (Table A1).}
In a second experiment we add
the \emph{alignment} (+align) and \emph{lemma features}
(+lemma) and show that this consistently outperforms all
baselines and edittree.  We then add the
\emph{dictionary feature} (+dict).  The resulting model
outperforms all previous models and is significantly better
than the best baselines for all languages.\footnote{We use the randomization test \cite{yeh2000} with $\alpha = .05$.}
These experiments show that \sw{lemming-p}
yields state-of-the-art results and that all our features are needed
to obtain optimal performance.
The improvements over the baselines are $>$1 for Czech and Latin and
$\geq$.5 for German and Hungarian.

\begin{table}
\renewcommand\thetable{1}
\centering
\small
\setlength{\tabcolsep}{1pt}
\begin{tabular}{l@{\hskip 4pt}lcccccc} \toprule
&         & cs & de & en & es & hu & la \\
\hline
\multirow{3}{*}{\begin{sideways}{\scriptsize baselines\,\,}\end{sideways}}
&{\scriptsize \software{simple}}      & 87.22\nig      & 93.27\nig      & 97.60\nig      & 92.92\nig      & 86.09\nig      & 85.19\nig \\
&{\scriptsize JCK}     & 96.24\nig      & \ul{97.67}\nig & \ul{98.71}\nig & 97.61\nig      & \ul{97.48}\nig & \ul{93.26}\nig \\
&{\scriptsize \sw{Morfette}} & \ul{96.25}\nig & 97.12\nig      & 98.43\nig      & \ul{97.97}\nig & 97.22\nig      & 91.89\nig \\
\midrule
\multirow{4}{*}{\begin{sideways}{\scriptsize \software{lemming-p}}\end{sideways}}
&{\scriptsize edittree} & 96.29\nig      & 97.84\sig      & 98.71\nig      & 97.91\nig      & 97.31\nig      & 93.00\nig\\
&{\scriptsize +align,+lemma} & 96.74\sig      & 98.17\sig      & 98.76\sig      & 98.05\nig      & 97.70\sig      & 93.76\sig\\
&{\scriptsize +dict} & \tb{97.50}\sig & \tb{98.36}\sig & \tb{98.84}\sig & 98.39\sig      & \tb{97.98}\sig & \tb{94.64}\sig\\
&{\scriptsize +mrph} & 96.59\sig      & 97.43\sig      &     \NA          & \tb{98.46}\sig & 97.77\sig      & 93.60\nig\\ \bottomrule
\end{tabular}
\caption{Lemma accuracy on \emph{dev} for the baselines and the
  different versions of \software{lemming-p}. POS
  and morphological attributes are predicted
  using \software{Morfette}. The best baseline numbers are underlined, the best numbers are bold.
  Models significantly better than the best baseline are marked (\sigsym).}
\tablabel{features}
\end{table}

The last experiment also uses the additional morphological
attributes predicted by \software{Morfette}
(+mrph). This leads to a drop in lemmatization
performance in all languages except Spanish (English has no additional attributes).
However, preliminary experiments showed that correct morphological
attributes would substantially improve lemmatization as they help in cases of ambiguity.
As an example,
number helps to lemmatize the singular German noun \word{Raps} \gloss{canola}, which looks like the plural of \word{Rap} \gloss{rap}.
Numbers can be found in Table A3 of the appendix.
This
motivates the necessity of \emph{joint tagging and lemmatization}.

For the final experiments, we run pipeline models on tags predicted by \marmot \cite{mueller2013}
and compare them to \software{Lemming-J}, 
the joint model described in
\secref{joint}. All \software{Lemming} versions use
exactly the same features.
\tabref{joint}
shows
that \software{Lemming-J} outperforms \software{Lemming-P}
in three
measures (see bold tag, lemma \& joint (tag+lemma) accuracies) except for English,
where we observe a tie in lemma accuracy and a small drop in
tag
and tag+lemma accuracy.
Coupling morphological
attributes and lemmatization (lines 8--10 vs 11--13) improves 
tag+lemma prediction for five languages. Improvements in lemma
accuracy of the joint over the best pipeline systems range
from .1 (Spanish), over $>$.3 (German, Hungarian) to
$\geq$.96 (Czech, Latin). 

Lemma accuracy improvements for
our best models (lines 4--13) over the best baseline (lines 2--3) are $>$1 (German, Spanish, Hungarian),  $>$2 (Czech, Latin)
and even more pronounced on unknown forms: $>$1 (English),
$>$5 (German, Spanish, Hungarian) and $>$12 (Czech, Latin).

\section{Conclusion}
\software{Lemming} is
a modular lemmatization model that
supports arbitrary global lemma features and joint modeling
of 
lemmata and 
morphological tags.
It is trainable on corpora annotated
with gold standard tags and lemmata, and does not rely on morphological
dictionaries or analyzers.
We 
have shown
that modeling lemmatization and tagging jointly
benefits both tasks,
and 
we
set the new state of the art in token-based lemmatization
on six languages.
\footnote{\cistern}\looseness=-1

\section*{Acknowledgments} 
We would like to thank the anonymous reviewers for their comments. 
The first author is a recipient of the Google Europe Fellowship in Natural Language Processing,
and this research is supported by this Google fellowship. 
The second author is supported by a Fulbright fellowship
awarded by the German-American Fulbright Commission
and the National Science Foundation under Grant No. 1423276.
This project has received funding from the European Union’s Horizon 2020 research and innovation
programme under grant agreement No 644402 (HimL) and the DFG grant
\emph{Models of Morphosyntax for Statistical Machine Translation}.
The fourth author was partially supported by
Deutsche Forschungsgemeinschaft (grant SCHU 2246/10-1).\looseness-1
\bibliographystyle{acl_natbib}
\bibliography{lemming}

\appendix
\onecolumn

\section{Tree Extraction}
\seclabel{ext}

\begin{algorithmic}[1]
  \Function{Tree}{$x$,$y$} 
  \State $i_s,i_e,j_s,j_e \gets \text{lcs}(x,y)$ 
  \If {$i_e - i_s = 0$}
  \State \Return \textsc{Sub}($x$,$y$)
  \Else 
  \State \Return (\textsc{Tree}($x_{0}^{i_s}, y_{0}^{j_s}$), $i_s$, \textsc{Tree}($x_{i_e}^{|x|}, y_{j_e}^{|y|}$), $|x| - i_e$)
  \EndIf
  \EndFunction
\end{algorithmic}

Create a tree given a form-lemma pair $\langle x,y \rangle$. \textsc{LCS} returns the start and end indexes of the LCS in $x$ and $y$. $x_{i_s}^{i_e}$ denotes the substring of $x$ starting at index $i_s$ (inclusive) and ending at index $i_e$ (exclusive). $i_e - i_s$ thus equals the length of this substring. $|x|$ denotes the length of $x$. Note that the tree does not store the LCS, but only the length of the prefix and suffix. This way the tree for \word{umgeschaut} can also be applied to transform \word{umgebaut} \gloss{renovated} into \word{umbauen} \gloss{to renovate}.

For the example \word{umgeschaut}-\word{umschauen}, the LCS is the stem \word{schau}. The function then recursively transforms \word{umge} into \word{um} and \word{t} into \word{en}.
The prefix and suffix lengths of the form are $4$ and $1$ respectively. The left sub-node needs to transform \word{umge} into \word{um}. The new LCS is \word{um}. The new prefix and suffix lengths are $0$ and $2$ respectively. As the new prefix is empty the is nothing more to do. The suffix node needs to transform \word{ge} into the empty string $\epsilon$. As the new LCS of the suffix is empty, because \word{ge} and $\epsilon$ have no character in common, the node is represented as a substitution node. The remaining transformation \word{t} into \word{en} is also represented as a substitution, resulting in the tree in \figref{edit-tree}:

\begin{figure}[!htb]
   \centering
    \includegraphics[width=0.40\textwidth]{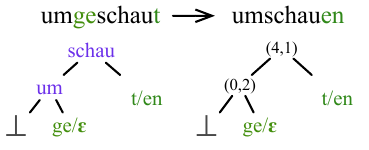}
\caption{Edit tree for the inflected form \word{umgeschaut} \gloss{looked around} 
and its lemma \word{umschauen} \gloss{to look around}. 
The right tree is the actual edit tree we use in our model,
the left tree visualizes what each node corresponds to.
Note
how the root node stores the length of the prefix \word{umge} and the suffix \word{t}.}
\figlabel{edit-tree}
\end{figure}

\section{Tree Application}
\seclabel{app}

\begin{algorithmic}[1]
  \Function{Apply}{$\tree$, $x$} 
  \If {$\tree$ is a \textsc{LCS} node}
  \State $\tree \rightarrow \tree_i, i_l, \tree_j, j_l $
  \If {$|x| < i_l + j_l$} \Comment{Prefix and Suffix do not fit.}
  \State \Return $\bot$
  \EndIf
  \State $p = $\textsc{Apply}$(\tree_i, x_0^{i_l})$ \Comment{Create prefix.}
  \If {$p$ is $\bot$} \Comment{Prefix tree cannot be applied.}
  \State \Return $\bot$
  \EndIf
  \State $s = $\textsc{Apply}$(\tree_j, x_{|x|-j_l}^{|x|})$ \Comment{Create suffix.}
  \If {$s$ is $\bot$} \Comment{Suffix tree cannot be applied.}
  \State \Return $\bot$
  \EndIf
  \State \Return p + $x_{i_l}^{|x|-j_l}$ + s \Comment{Concatenate prefix, \textsc{LCS} and suffix.}
  \Else \Comment{$\tree$ is a \textsc{Sub} node} 
  \State $\tree \rightarrow u, v$
  \If {$x = u$} \Comment{If $x$ and $u$ match return $v$}
  \State \Return v
  \EndIf
  \State \Return $\bot$ \Comment{$\tree$ cannot be applied.}
  \EndIf
  \EndFunction
\end{algorithmic}

In the code + represents string concatenation and $\bot$ a null
string, meaning that the tree cannot be applied to the form.  We first
run the tree depicted in \figref{edit-tree} on the form
\word{angebaut} \gloss{attached (to a building)}.  The first node is a
\textsc{LCS} node specifying that prefix and suffix should have length
$4$ and $1$, respectively.  We thus recursively apply the left child
node to the prefix string \word{ange}. This is done by matching the
length two prefix \word{an} and deleting \word{ge} yielding the
intermediate result \word{anbaut}. We continue on the right side of
the tree and replace \word{t} with \word{en}. This yield the final
(correct) result \word{anbauen}.  The application of the tree to the
form \word{einbauen} \gloss{installed} would fail, as we would try to
substitute a \word{ge} in \word{eing}.

\vspace{-12pt}
\section{Development Results}
\vspace{-12pt}
\seclabel{dev}
\begin{table}[H]
\renewcommand\thetable{A1}
\centering
\small
\setlength{\tabcolsep}{2pt}
\begin{tabular}{l@{\hskip 4pt}llrrrrrrrrrrrr}
&&& \multicolumn{2}{c}{cs} & \multicolumn{2}{c}{de} & \multicolumn{2}{c}{en} & \multicolumn{2}{c}{es} & \multicolumn{2}{c}{hu} & \multicolumn{2}{c}{la} \\
\hline
\multirow{15}{*}{\begin{sideways}{\Large Baselines}\end{sideways}}
& \sw{simple} & tag & 86.08 & 69.83 & 79.05 & 56.02 & 94.72 & 87.95 & 96.29 & 87.21 & 94.37 & 85.20 & 83.86 & 57.77 \\
& \sw{morfette} & lemma & 87.22 & 32.95 & 93.27 & 62.15 & 97.60 & 75.64 & 92.92 & 59.68 & 86.09 & 42.02 & 85.19 & 14.06 \\
&  & joint & 77.18 & 22.56 & 75.60 & 36.43 & 93.15 & 67.58 & 90.28 & 54.00 & 82.99 & 37.03 & 75.00 & 5.39 \\
\cline{2-15}
& \sw{simple} & tag & \ul{89.82} & \ul{76.83} & \ul{85.05} & \ul{65.22} & \ul{\tb{95.71}} & \ul{\tb{90.29}} & \ul{96.83} & \ul{89.00} & \ul{\tb{95.46}} & \ul{88.17} & \ul{86.35} & \ul{65.21} \\
& \marmot & lemma & 87.36 & 32.95 & 93.28 & 62.15 & 97.66 & 75.64 & 92.99 & 59.68 & 86.11 & 42.02 & 85.35 & 14.06 \\
&  & joint & 79.99 & 23.98 & 80.71 & 40.26 & 94.09 & 69.48 & 90.75 & 54.66 & 83.47 & 37.16 & 76.58 & 6.61 \\
\cline{2-15}
& JCK & tag & 86.08 & 69.83 & 79.05 & 56.02 & 94.72 & 87.95 & 96.29 & 87.21 & 94.37 & 85.20 & 83.86 & 57.77 \\
& \sw{morfette} & lemma & 96.24 & 82.59 & 97.67 & 88.80 & 98.71 & 92.50 & 97.61 & 86.76 & 97.48 & 91.16 & \ul{93.26} & \ul{63.09} \\
&  & joint & 84.32 & 61.67 & 78.10 & 51.42 & 94.43 & 84.61 & 94.67 & 78.39 & 93.15 & 80.73 & 81.34 & 43.71 \\
\cline{2-15}
& JCK & tag & \ul{89.82} & \ul{76.83} & \ul{85.05} & \ul{65.22} & \ul{\tb{95.71}} & \ul{\tb{90.29}} & \ul{96.83} & \ul{89.00} & \ul{\tb{95.46}} & \ul{88.17} & \ul{86.35} & \ul{65.21} \\
& \marmot & lemma & \ul{96.26} & \ul{82.88} & \ul{97.74} & 89.11 & \ul{98.81} & \ul{93.06} & 97.78 & 87.10 & \ul{97.68} & \ul{91.91} & 93.06 & 62.64 \\
&  & joint & \ul{88.17} & \ul{68.43} & \ul{84.16} & \ul{60.39} & \ul{95.41} & \ul{86.95} & \ul{95.37} & 80.25 & \ul{94.33} & \ul{83.70} & \ul{83.57} & \ul{48.84} \\
\cline{2-15}
& \sw{morfette} & tag & 86.08 & 69.83 & 79.05 & 56.02 & 94.72 & 87.95 & 96.29 & 87.21 & 94.37 & 85.20 & 83.86 & 57.77 \\
& \sw{morfette} & lemma & 96.25 & 82.54 & 97.12 & \ul{89.90} & 98.43 & 92.54 & \ul{97.97} & \ul{89.94} & 97.22 & 90.04 & 91.89 & 55.13 \\
&  & joint & 84.39 & 61.94 & 77.60 & 52.87 & 94.16 & 84.70 & 95.09 & \ul{81.95} & 93.11 & 80.60 & 80.09 & 36.39 \\
\hline
\hline
\multirow{27}{*}{\begin{sideways}{\Large \software{lemming}}\end{sideways}}
& edit tree & tag & 86.08 & 69.83 & 79.05 & 56.02 & 94.72 & 87.95 & 96.29 & 87.21 & 94.37 & 85.20 & 83.86 & 57.77 \\
& \sw{morfette} & lemma & 96.29 & 82.93 & 97.84 & 89.78 & 98.71 & 92.63 & 97.91 & 89.00 & 97.31 & 90.49 & 93.00 & 61.68 \\
&  & joint & 84.45 & 62.36 & 78.32 & 52.75 & 94.43 & 84.79 & 94.97 & 80.68 & 93.08 & 80.48 & 80.92 & 41.27 \\
\cline{2-15}
& align & tag & 86.08 & 69.83 & 79.05 & 56.02 & 94.72 & 87.95 & 96.29 & 87.21 & 94.37 & 85.20 & 83.86 & 57.77 \\
& \sw{morfette} & lemma & 96.74 & 85.38 & 98.17 & 91.61 & 98.76 & 93.11 & 98.05 & 90.13 & 97.70 & 92.15 & 93.76 & 66.30 \\
&  & joint & 84.72 & 63.88 & 78.47 & 53.52 & 94.47 & 85.09 & 95.05 & 81.29 & 93.31 & 81.51 & 81.49 & 44.67 \\
\cline{2-15}
& dict & tag & 86.08 & 69.83 & 79.05 & 56.02 & 94.72 & 87.95 & 96.29 & 87.21 & 94.37 & 85.20 & 83.86 & 57.77 \\
& \sw{morfette} & lemma & 97.50 & 89.38 & 98.36 & 92.66 & 98.84 & 94.02 & 98.39 & 92.56 & 97.98 & 93.30 & 94.64 & 71.57 \\
&  & joint & 85.06 & 65.60 & 78.53 & 53.91 & 94.53 & 85.83 & 95.27 & 82.89 & 93.46 & 82.10 & 81.87 & 46.92 \\
\cline{2-15}
& morph & tag & 86.08 & 69.83 & 79.05 & 56.02 & \NA & \NA & 96.29 & 87.21 & 94.37 & 85.20 & 83.86 & 57.77 \\
& \sw{morfette} & lemma & 96.59 & 87.28 & 97.43 & 89.96 & \NA & \NA & 98.46 & 92.98 & 97.77 & 92.62 & 93.60 & 66.30 \\
&  & joint & 85.64 & 67.73 & 78.87 & 55.17 & \NA & \NA & 95.63 & 84.69 & 94.10 & 84.02 & 82.29 & 48.72 \\
\cline{2-15}
& edit tree & tag & 89.82 & 76.83 & 85.05 & 65.22 & \tb{95.71} & \tb{90.29} & 96.83 & 89.00 & \tb{95.46} & 88.17 & 86.35 & 65.21 \\
& \marmot & lemma & 96.33 & 83.04 & 97.93 & 90.05 & 98.82 & 93.02 & 98.14 & 89.61 & 97.53 & 91.22 & 92.85 & 60.98 \\
&  & joint & 88.23 & 68.85 & 84.41 & 61.89 & 95.40 & 87.00 & 95.75 & 83.06 & 94.30 & 83.58 & 83.20 & 46.73 \\
\cline{2-15}
& align & tag & 89.82 & 76.83 & 85.05 & 65.22 & \tb{95.71} & \tb{90.29} & 96.83 & 89.00 & \tb{95.46} & 88.17 & 86.35 & 65.21 \\
& \marmot & lemma & 96.80 & 85.72 & 98.24 & 91.94 & 98.88 & 93.71 & 98.27 & 90.75 & 97.91 & 92.91 & 93.50 & 64.89 \\
&  & joint & 88.57 & 70.69 & 84.59 & 62.95 & 95.45 & 87.47 & 95.81 & 83.64 & 94.52 & 84.60 & 83.83 & 50.32 \\
\cline{2-15}
& dict & tag & 89.82 & 76.83 & 85.05 & 65.22 & \tb{95.71} & \tb{90.29} & 96.83 & 89.00 & \tb{95.46} & 88.17 & 86.35 & 65.21 \\
& \marmot & lemma & 97.62 & 89.88 & 98.43 & 92.95 & 98.94 & \tb{94.28} & 98.62 & 93.23 & 98.23 & 94.16 & 94.49 & 70.60 \\
&  & joint & 88.97 & 72.79 & 84.63 & 63.16 & \tb{95.50} & \tb{88.08} & 96.02 & 85.24 & 94.73 & 85.46 & 84.41 & 53.79 \\
\cline{2-15}
& morph & tag & 89.82 & 76.83 & 85.05 & 65.22 & \NA & \NA & 96.83 & 89.00 & \tb{95.46} & 88.17 & 86.35 & 65.21 \\
& \marmot & lemma & 97.38 & 89.55 & 98.24 & 92.42 & \NA & \NA & \tb{98.71} & 93.97 & 98.30 & 94.38 & 94.39 & 70.54 \\
&  & joint & 89.30 & 74.32 & 84.81 & 64.08 & \NA & \NA & 96.17 & 86.29 & 95.15 & 86.87 & 84.62 & 55.07 \\
\cline{2-15}
& dict & tag & \tb{90.47} & 78.37 & 85.31 & 65.94 & 95.66 & 88.60 & 96.94 & 89.25 & 95.29 & 87.55 & \tb{86.77} & 65.98 \\
& joint & lemma & 98.34 & 92.96 & 98.66 & 94.07 & \tb{99.00} & 94.23 & 98.68 & 94.08 & 98.53 & 95.46 & \tb{96.68} & \tb{82.99} \\
&  & joint & 89.82 & 75.47 & 84.94 & 64.19 & 95.48 & 86.52 & 96.11 & 85.43 & 94.71 & 85.40 & 85.85 & 60.98 \\
\cline{2-15}
& morph & tag & 90.35 & \tb{79.61} & \tb{85.52} & \tb{67.69} & 95.66 & 88.60 & \tb{96.95} & \tb{89.55} & \tb{95.46} & \tb{88.85} & 86.70 & \tb{66.82} \\
& joint & lemma & \tb{98.55} & \tb{94.12} & \tb{98.74} & \tb{94.47} & \tb{99.00} & 94.23 & 98.68 & \tb{94.53} & \tb{98.57} & \tb{95.65} & 96.51 & 82.22 \\
&  & joint & \tb{90.05} & \tb{78.37} & \tb{85.28} & \tb{66.58} & 95.48 & 86.52 & \tb{96.20} & \tb{86.51} & \tb{95.30} & \tb{88.22} & \tb{85.97} & \tb{62.97} \\
\end{tabular}
 \caption{{\small Development accuracies for the baselines and the different pipeline versions and a joint version of \software{lemming}. The numbers in each cell are general token accuracy and the token accuracy on unknown forms. Each cell specifies either a baseline $\in \{$baseline, JCK, \sw{Morfette}$\}$ or a \sw{Lemming} feature set $\in \{$edit tree, align, dict, morph$\}$ and a tagger $\in \{$\sw{Morfette}, \marmot, joint$\}$.}}
\vspace{-12pt}
\end{table}

\vspace{-12pt}
\section{Test Results}
\vspace{-12pt}
\seclabel{tst}
\begin{table}[H]
\renewcommand\thetable{A2}
\centering
\small
\setlength{\tabcolsep}{2pt}
\begin{tabular}{l@{\hskip 4pt}llrrrrrrrrrrrr}
&&& \multicolumn{2}{c}{cs} & \multicolumn{2}{c}{de} & \multicolumn{2}{c}{en} & \multicolumn{2}{c}{es} & \multicolumn{2}{c}{hu} & \multicolumn{2}{c}{la} \\
\hline
\multirow{15}{*}{\begin{sideways}{\Large Baselines}\end{sideways}}
& \sw{simple} & tag & 86.00 & 68.88 & 76.80 & 52.86 & 95.35 & 87.58 & 96.30 & 87.65 & 92.33 & 81.32 & 79.70 & 47.06 \\
& \sw{morfette} & lemma & 86.81 & 31.16 & 91.42 & 58.68 & 97.88 & 78.95 & 92.86 & 59.34 & 84.67 & 38.95 & 83.60 & 20.43 \\
&  & joint & 76.96 & 21.22 & 72.83 & 33.27 & 93.82 & 68.98 & 90.14 & 53.45 & 80.41 & 33.08 & 71.62 & 7.52 \\
\cline{2-15}
& \sw{simple} & tag & \ul{89.75} & \ul{76.83} & \ul{82.81} & \ul{61.60} & \ul{\tb{96.45}} & \ul{\tb{90.68}} & \ul{97.05} & \ul{90.07} & \ul{93.64} & \ul{84.65} & \ul{82.37} & \ul{53.73} \\
& \marmot & lemma & 86.94 & 31.16 & 91.48 & 58.68 & 97.92 & 78.95 & 92.92 & 59.34 & 84.73 & 38.95 & 83.80 & 20.43 \\
&  & joint & 79.66 & 22.86 & 77.78 & 36.49 & 94.87 & 71.62 & 90.81 & 54.98 & 81.06 & 33.27 & 73.51 & 8.58 \\
\cline{2-15}
& JCK & tag & 86.00 & 68.88 & 76.80 & 52.86 & 95.35 & 87.58 & 96.30 & 87.65 & 92.33 & 81.32 & 79.70 & 47.06 \\
& \sw{morfette} & lemma & 95.87 & 80.79 & 96.50 & 85.54 & 98.95 & 93.76 & 97.58 & 86.80 & 96.52 & 88.09 & \ul{90.84} & 58.13 \\
&  & joint & 84.00 & 59.46 & 75.60 & 47.77 & 95.09 & 84.34 & 94.54 & 77.97 & 90.71 & 75.62 & 76.76 & 33.56 \\
\cline{2-15}
& JCK & tag & \ul{89.75} & \ul{76.83} & \ul{82.81} & \ul{61.60} & \ul{\tb{96.45}} & \ul{\tb{90.68}} & \ul{97.05} & \ul{90.07} & \ul{93.64} & \ul{84.65} & \ul{82.37} & \ul{53.73} \\
& \marmot & lemma & \ul{95.95} & \ul{81.28} & \ul{96.63} & 85.84 & \ul{99.08} & \ul{94.28} & 97.69 & 87.19 & \ul{96.69} & \ul{88.66} & 90.79 & \ul{58.23} \\
&  & joint & \ul{87.85} & \ul{67.00} & \ul{81.60} & \ul{55.97} & \ul{96.17} & \ul{87.32} & \ul{95.44} & 80.62 & \ul{92.15} & \ul{78.89} & \ul{79.51} & \ul{39.07} \\
\cline{2-15}
& \sw{morfette} & tag & 86.00 & 68.88 & 76.80 & 52.86 & 95.35 & 87.58 & 96.30 & 87.65 & 92.33 & 81.32 & 79.70 & 47.06 \\
& \sw{morfette} & lemma & 95.88 & 80.70 & 95.88 & \ul{87.55} & 98.73 & 93.56 & \ul{98.02} & \ul{90.16} & 96.06 & 86.21 & 90.08 & 54.16 \\
&  & joint & 84.10 & 59.92 & 74.87 & 49.22 & 94.89 & 84.44 & 95.03 & \ul{81.68} & 90.60 & 75.11 & 76.57 & 32.50 \\
\hline
\hline
\multirow{27}{*}{\begin{sideways}{\Large \software{lemming}}\end{sideways}}
& edit tree & tag & 86.00 & 68.88 & 76.80 & 52.86 & 95.35 & 87.58 & 96.30 & 87.65 & 92.33 & 81.32 & 79.70 & 47.06 \\
& \sw{morfette} & lemma & 95.91 & 81.07 & 96.82 & 87.31 & 98.95 & 93.69 & 97.87 & 88.97 & 96.15 & 86.68 & 91.20 & 59.87 \\
&  & joint & 84.15 & 60.27 & 75.86 & 49.22 & 95.08 & 84.37 & 94.86 & 80.38 & 90.50 & 74.86 & 77.00 & 34.73 \\
\cline{2-15}
& align & tag & 86.00 & 68.88 & 76.80 & 52.86 & 95.35 & 87.58 & 96.30 & 87.65 & 92.33 & 81.32 & 79.70 & 47.06 \\
& \sw{morfette} & lemma & 96.43 & 83.83 & 97.22 & 89.25 & 98.99 & 94.15 & 98.04 & 90.28 & 96.87 & 89.72 & 91.64 & 62.36 \\
&  & joint & 84.47 & 61.97 & 76.05 & 50.07 & 95.11 & 84.73 & 94.98 & 81.30 & 90.94 & 76.76 & 77.18 & 35.79 \\
\cline{2-15}
& dict & tag & 86.00 & 68.88 & 76.80 & 52.86 & 95.35 & 87.58 & 96.30 & 87.65 & 92.33 & 81.32 & 79.70 & 47.06 \\
& \sw{morfette} & lemma & 97.24 & 88.13 & 97.58 & 91.03 & 99.07 & 95.06 & 98.32 & 92.36 & 97.28 & 91.32 & 92.97 & 69.14 \\
&  & joint & 84.88 & 64.14 & 76.13 & 50.48 & 95.19 & 85.55 & 95.19 & 82.91 & 91.14 & 77.48 & 77.52 & 37.59 \\
\cline{2-15}
& morph & tag & 86.00 & 68.88 & 76.80 & 52.86 & \NA & \NA & 96.30 & 87.65 & 92.33 & 81.32 & 79.70 & 47.06 \\
& \sw{morfette} & lemma & 96.52 & 86.69 & 96.57 & 88.26 & \NA & \NA & 98.53 & 93.73 & 96.70 & 89.30 & 91.53 & 62.31 \\
&  & joint & 85.57 & 66.73 & 76.56 & 51.90 & \NA & \NA & 95.63 & 85.02 & 91.92 & 79.73 & 78.18 & 39.65 \\
\cline{2-15}
& edit tree & tag & 89.75 & 76.83 & 82.81 & 61.60 & \tb{96.45} & \tb{90.68} & 97.05 & 90.07 & 93.64 & 84.65 & 82.37 & 53.73 \\
& \marmot & lemma & 96.05 & 81.68 & 96.99 & 87.69 & 99.08 & 94.25 & 98.00 & 89.56 & 96.38 & 87.54 & 91.37 & 60.88 \\
&  & joint & 88.03 & 68.04 & 81.96 & 57.91 & 96.17 & 87.38 & 95.80 & 83.42 & 91.93 & 78.15 & 79.75 & 40.34 \\
\cline{2-15}
& align & tag & 89.75 & 76.83 & 82.81 & 61.60 & \tb{96.45} & \tb{90.68} & 97.05 & 90.07 & 93.64 & 84.65 & 82.37 & 53.73 \\
& \marmot & lemma & 96.56 & 84.49 & 97.38 & 89.68 & 99.12 & 94.51 & 98.17 & 90.89 & 97.11 & 90.61 & 91.80 & 63.26 \\
&  & joint & 88.41 & 70.03 & 82.20 & 59.02 & 96.19 & 87.61 & 95.94 & 84.38 & 92.41 & 80.14 & 80.03 & 41.87 \\
\cline{2-15}
& dict & tag & 89.75 & 76.83 & 82.81 & 61.60 & \tb{96.45} & \tb{90.68} & 97.05 & 90.07 & 93.64 & 84.65 & 82.37 & 53.73 \\
& \marmot & lemma & 97.46 & 89.14 & 97.70 & 91.27 & \tb{99.21} & \tb{95.59} & 98.48 & 92.98 & 97.53 & 92.10 & 93.07 & 69.83 \\
&  & joint & 88.86 & 72.51 & 82.27 & 59.42 & \tb{96.27} & \tb{88.49} & 96.12 & 85.80 & 92.59 & 80.77 & 80.49 & 44.26 \\
\cline{2-15}
& morph & tag & 89.75 & 76.83 & 82.81 & 61.60 & \NA & \NA & 97.05 & 90.07 & 93.64 & 84.65 & 82.37 & 53.73 \\
& \marmot & lemma & 97.29 & 88.98 & 97.51 & 90.85 & \NA & \NA & 98.68 & 94.32 & 97.53 & 92.15 & 92.54 & 67.81 \\
&  & joint & 89.23 & 74.24 & 82.49 & 60.42 & \NA & \NA & 96.35 & 87.25 & 93.11 & 82.56 & 80.67 & 45.21 \\
\cline{2-15}
& dict & tag & \tb{90.34} & 78.47 & \tb{83.10} & 62.36 & 96.32 & 89.70 & 97.11 & 90.13 & 93.64 & 84.78 & 82.89 & 54.69 \\
& joint & lemma & 98.27 & 92.67 & \tb{98.10} & 92.79 & \tb{99.21} & 95.23 & 98.67 & 94.07 & 98.02 & 94.15 & \tb{95.58} & \tb{81.74} \\
&  & joint & 89.69 & 75.44 & 82.64 & 60.49 & 96.17 & 87.87 & 96.23 & 86.19 & 92.84 & 81.89 & 81.92 & 49.97 \\
\cline{2-15}
& morph & tag & 90.20 & \tb{79.72} & \tb{83.10} & \tb{63.10} & 96.32 & 89.70 & \tb{97.17} & \tb{90.66} & \tb{93.67} & \tb{85.12} & \tb{83.49} & \tb{58.76} \\
& joint & lemma & \tb{98.42} & \tb{93.46} & \tb{98.10} & \tb{93.02} & \tb{99.21} & 95.23 & \tb{98.78} & \tb{94.86} & \tb{98.08} & \tb{94.26} & 95.36 & 80.94 \\
&  & joint & \tb{89.90} & \tb{78.34} & \tb{82.84} & \tb{62.10} & 96.17 & 87.87 & \tb{96.41} & \tb{87.47} & \tb{93.40} & \tb{84.15} & \tb{82.57} & \tb{54.63} \\
\end{tabular}
 \caption{{\small Test accuracies for the baselines and the different pipeline versions and a joint version of \software{lemming}. The numbers in each cell are general token accuracy and the token accuracy on unknown forms. Each cell specifies either a baseline $\in \{$baseline, JCK, \sw{Morfette}$\}$ or a \sw{Lemming} feature set $\in \{$edit tree, align, dict, morph$\}$ and a tagger $\in \{$\sw{Morfette}, \marmot, joint$\}$.}}
\vspace{-12pt}
\end{table}

\newpage

\begin{table}[h]
\renewcommand\thetable{A3}
\centering
\small
\setlength{\tabcolsep}{3pt}
\begin{tabular}{l@{\hskip 4pt}lccccc}
         & cs        & de    & es    & hu & la\\
\hline
{\scriptsize +dict } & 98.35     & 99.04 & 98.92 & 98.83 & 96.14\\
{\scriptsize +morph} & \tb{99.03}& \tb{99.47} & \tb{99.09} & \tb{99.41} & \tb{96.73}\\
\end{tabular}
\caption{Development accuracies for \sw{Lemming} with and without
morphological attributes using {\em gold} tags.}
\tablabel{gold}
\end{table}

\begin{table}[h]
\renewcommand\thetable{A4}
\centering
\small
\setlength{\tabcolsep}{3pt}
\begin{tabular}{c|rrrr|rrrr|rrrr}
   & \multicolumn{4}{|c}{train} & \multicolumn{4}{|c}{dev} & \multicolumn{4}{|c}{test}\\
   & sent & token &  pos & morph & sent & token  & form unk & lemma unk & sent & token  & form unk & lemma unk\\
\hline
cs &  5979 & 100012 &   12 &  266 &  5228 &  87988 & 19.86 &  9.79 &  4213 &  70348 & 19.89 &  9.73 \\
de &  5662 & 100009 &   51 &  204 &  5000 &  76704 & 17.11 & 13.53 &  5000 &  92004 & 19.51 & 15.64 \\
en &  4028 & 100012 &   46 &      &  1336 &  32092 &  8.06 &  6.16 &  1640 &  39590 &  8.52 &  6.56 \\
es &  3431 & 100027 &   12 &  226 &  1655 &  50368 & 13.37 &  8.85 &  1725 &  50630 & 13.41 &  9.16 \\
hu &  4390 & 100014 &   22 &  572 &  1051 &  29989 & 23.51 & 14.36 &  1009 &  19908 & 24.80 & 14.64 \\
la &  7122 &  59992 &   23 &  474 &   890 &   9475 & 17.59 &  6.43 &   891 &   9922 & 19.82 &  7.56 \\
\end{tabular}
\caption{Dataset statistics. Showing number of sentences (sent), tokens (token), POS tags (pos), morphological tags (morph) and token-based unknown form (form unk) and lemma (lemma unk) rates.}
\end{table}

\end{document}